\documentclass[runningheads]{llncs}
\usepackage[table,xcdraw]{xcolor}
\usepackage{caption}
\usepackage{subcaption}
\usepackage{graphicx}
\usepackage{hyperref}
\usepackage{todonotes}

\usepackage{multirow}

\begin{document}
\title{Dialog speech sentiment classification for imbalanced datasets.}
%
%
\author{Sergis Nicolaou \inst{1}\and
Lambros Mavrides  \inst{1} \and
Georgina Tryfou \inst{1}  \and
Kyriakos Tolias  \inst{2} \and Konstantinos Panousis  \inst{2} \and Sotirios Chatzis  \inst{2} \and
Sergios Theodoridis  \inst{3}
}

\authorrunning{S. Nicolaou et al.}
%
\institute{AI Team, Impactech LTD, Limassol, Cyprus \\
\email{\{s.nicolaou,l.mavrides,g.tryfou\}@impactechs.com} \\
\url{ http://ai.impactechs.com} \and
Cyprus University of Technology (CUT) \\ 
\email{\{kv.tolias,k.panousis, sotirios.chatzis\}@cut.ac.cy}
\and
Aalborg University, Denmark  }
\maketitle              
\begin{abstract}
Speech is the most common way humans express their feelings, and sentiment analysis is the use of tools such as natural language processing and computational algorithms to identify the polarity of these feelings. Even though this field has seen tremendous advancements in the last two decades, the task of effectively detecting under represented sentiments in different kinds of datasets is still a challenging task. In this paper, we use single and bi-modal analysis of short dialog utterances and gain insights on the main factors that aid in sentiment detection, particularly in the underrepresented classes, in datasets with and without inherent sentiment component. Furthermore, we propose an architecture which uses a learning rate scheduler and different monitoring criteria and provides state-of-the-art results for the SWITCHBOARD imbalanced sentiment dataset. 


\end{abstract}
\section{Introduction}
In human communication, sentiment is liaised via posture and facial expressions as well as via speech. 
In this context, sentiment analysis is the task of classifying a segment of spoken text, for example a dialog turn, into a class that better describes the speaker's state of mind, such as positive, neutral or negative.
A dialog-based sentiment analysis system can be used for marketing purposes, and for monitoring and optimizing the performance of agents in sales calls \cite{sentometrics}.
In a different use case, people with a hearing disability can immensely benefit from a machine that can understand and convey human communication. 
In order to develop such machines, it is paramount that they are able to understand sentiment, with a focus on perceiving subtle positive and negative sentiment cues. 

Despite the significant role that sentiment analysis plays in different intelligent applications, and the extensive use of modern deep learning methods to address it, it is still a challenging task \cite{shayaa2018sentiment}. 
The first reason for this is the variance in the way that humans may choose to express sentiment when they speak. They can use the tone of their voice, words or other subtle cues in order to express in a very controlled way strong sentiments.   
Second, there are still not enough annotated and large corpora that can be publicly found and used for the development of speech sentiment analysis systems. Therefore, systems are often prone to generalization problems.
On top of this, the quality of sentiment found in some of the most commonly used speech corpora in the field, is highly affected by the type and original purpose of the corpus \cite{hussein2018survey}.
Therefore, there is still a need to investigate speech sentiment analysis, with a focus on the factors that can contribute to improving results for any dataset, and with an emphasis in the underrepresented classes which are often more prone to errors.

In this paper, we investigate in detail the effect of using single modality and  bi-modal speech sentiment analysis, \textit{i.e.} sentiment analysis performed simultaneously on speech and text.
First, we propose our in-house single modal classification approaches for both modalities, and investigate their behaviour in all classes. We perform a detailed study on how we can improve the behaviour of the single modal classifiers in underrepresented classes, which may often be more important for certain applications.
Finally, we investigate how the effects that have been identified in the single modal classification are generalized in the bi-modal scenario. 


The remainder of this paper is organized as follows. In section \ref{sec:rel_work}, we outline the current state-of-the-art in the area of speech sentiment analysis. 
In section \ref{sec:data}, we present the datasets on which we base our research and quickly outline their main characteristics. 
In section \ref{sec:exp_results}, we review the experimental setup, present the corresponding experiments and discuss the results. Finally, conclusions are drawn and future steps are described in section \ref{sec:conclusions}.

\section{Related Work} \label{sec:rel_work}

The first attempts at sentiment analysis involved the use of textual data, as this kind of data was easier to find and process. Previous work on textual sentiment analysis is outlined in \cite{shi2019survey}, and it is evident that since the latter half of the 2010s, more and more emphasis was put on deep learning approaches, such as convolutional neural networks (CNN) and recurrent neural networks (RNN). Moreover, many attempts have been made to create hybrid models which combine deep learning methods with each other, or with traditional machine learning approaches like SVM and lexicon-based methods. In addition, the use of attention layers, \cite{vaswani2017attention}, has become increasingly popular for solving sentiment analysis tasks. The current state-of-the-art results in popular datasets such as Amazon have been achieved using architectures that incorporate attention \cite{xie2019unsupervised}. 

As other forms of data, such as audio and video, became more readily available in recent years, their use in sentiment analysis tasks has also increased. In particular, acoustic data from audio has the potential of being a very useful tool in sentiment analysis. This is due to the nature of human speech and the ability of human to convey sentiment using their voice. Various studies such as \cite{laukka2013cross,nordstrom2017emotion} have established that language-independent vocalisations are rich in emotional content and information like sentiment can be conveyed across cultures. These results indicate that acoustic features can be an extremely powerful tool for sentiment analysis.
Sentiment analysis using acoustic features has not received as much in-depth research as textual sentiment analysis, and much of the related work is focused in speech emotion recognition rather than sentiment analysis.

Lately, a lot of research is performed in multi-modal systems for sentiment analysis, where features from different input types, such as text, audio and video are processed simultaneously to improve performance. 
\cite{li2019dilated} experimented with residual networks and attention on the IEMOCAP dataset \cite{busso2008iemocap} and achieved an accuracy of 67.4\%. On the same dataset, \cite{poria2018multimodal} reported an accuracy of 66.2 \% using a bidirectional LSTM architecture.

In \cite{kim2019dnn}, acoustic features are fused with lexical features at an early stage, and the fused features are fed to a classification DNN for utterance level emotion classification. The authors reported a classification accuracy of 75.5\% on IEMOCAP.
In \cite{cho2019deep}, acoustic features are processed with an LSTM network that predicts emotions and word sequences text features are fed to a multi-resolution CNN trained for the same task. The prefinal outputs of the two are combined with an SVM to produce the final classification verdict. The authors report a significant improvement in the IEMOCAP dataset, but the finding is not confirmed in a proprietary telephone speech dataset that they use. 
In \cite{gu2018multimodal}, the authors propose a hierarchical bi-modal architecture with attention and word-level fusion to classify utterance-level sentiment and emotion from text and audio data. 
A different approach to fuse information from various modalities is presented in \cite{lu2020speech}, where the authors propose to use pre-trained ASR features and solve the sentiment analysis as a down-stream task. With the assumption that ASR features encode acoustic as well as linguistic information the authors achieve state-of-the-art results. Specifically, they achieved an accuracy of 71.7\% on the IEMOCAP dataset and 70.1\% on the SWITCHBOARD dataset.

Although bi-modal sentiment analysis from speech and text is not a new domain there is not enough research on how the different modalities perform in detecting the various sentiments, and to which extend each modality affects the reported results. Also, the use of different datasets, often with very different class weights, does not allow one to clearly attribute the success of a certain system either in detecting the most popular class or a good prediction of all classes. Fully comprehending the reported results becomes even more challenging as the two modalities may result in contradicting numbers. 

\section{Data} \label{sec:data}

\subsection{SWITCHBOARD-sentiment dataset}
The SWITCHBOARD-1 Telephone Speech Corpus\footnote{https://catalog.ldc.upenn.edu/LDC97S62} is a large speech dataset very commonly used for training and bench-marking ASR systems. A subset of the SWITCHBOARD dataset was annotated with sentiment labels, released as SWBDsentiment dataset \footnote{https://catalog.ldc.upenn.edu/LDC2020T14}. It consists of 3 sentiment labels (positive, negative, and neutral) for approximately 49,500 utterances covering 140 hours of SWITCHBOARD audio.
For each segment we selected the sentiment which voted by the majority of the annotators, while discarding the remaining. After this process, the final data constist of
25445 neutral, 15308 positive and 8549 negative examples. Hence the negative class is the minority class in this dataset.
For our experiments, we perform stratified 10-fold cross validation. Segments from one dialog are kept in a single fold so as not to bias the model \cite{chen2020large,lu2020speech}. Although the original purpose of the dataset creation was not related to the sentiment analysis task, the speakers did receive a direction to speak as natural as possible and converse in a realistic way. This, along with the wide variety of topics covered in the dataset makes the SWITCHBOARD a valuable resource in the study of bi-modal sentiment analysis


\subsection{IEMOCAP-3 dataset}
The IEMOCAP dataset \cite{busso2008iemocap} consists of dyadic interactions between actors. There are five sessions, each with a male and a female
speaker, for 10 unique speakers total. As commonly performed in the literature, we consider only utterances with majority agreement ground-truth labels. 
To create the version we are using, which we call IEMOCAP-3, we create three sentiment categories, namely the neutral, positive and negative classes. To create the positive class we combine the happy and excited labels, and to create the negative we combine the angry and sad labels. Following the removal of utterances which contained only non-verbal actions such as breathing and laughing, the dataset contains 4156 negative, 1703 neutral and 1709 positive utterances. In this dataset the negative class is the majority class as it is about 3 times bigger than the other two classes. It is noted here that in the context of sentiment analysis, IEMOCAP is commonly split in the literature into four class, however we believe that for the scope of our research creating three classes is a better fit.
For training, we perform the same split approach described above (stratified 10 fold cross validation, without dialog overlap among folds).

\section {Experiments and results} \label{sec:exp_results}

\subsection{Acoustic sentiment analysis}
For acoustic sentiment analysis, our data processing starts with the extraction of acoustic features, using the Librosa python toolkit. We experimented with various features such as spectrograms, MFCCs with their first and second order derivatives, pitch and chroma features. We observed similar behaviour for most of these features. In the following experiments we report results from MFCCs as these showed a slight improvement in the classification accuracy for the under represented classes. 
In order to get the same number of MFCC frames per audio, even though they are of various sizes, we use a variable window length and a 25\% overlap so we always get 300 MFCC vectors per audio. This processing approach results in the same time resolution between the different audio files, but the time axis is either stretched or compressed, to ensure the same number of MFCC vectors per audio. The idea is similar to other acoustic data augmentation techniques used in the literature, as for example time warping discussed in \cite{park2019specaugment}.
Each MFCC vector contains 20 coefficients, and is then augmented with its first and second order derivatives, resulting to a 300x60 feature matrix per signal.

\begin{figure}
     \centering
  
         \includegraphics[width=\textwidth]{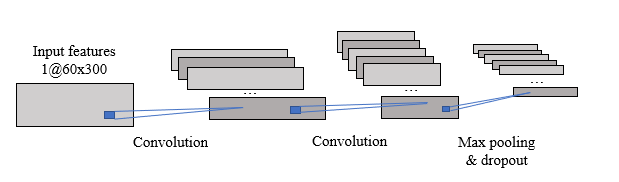}
         \caption{The convolutional block that we use as a building block for the CNN.}
        \label{fig:conv_block}
\end{figure}

We train a CNN using the above features extracted from the audio. Our architecture consists of consecutive convolutional blocks, each containing two 2D-convolutional layers with batch normalization and ReLU activation, followed by a max pooling and a dropout layer, as shown in figure \ref{fig:conv_block}.
Since we address sentiment analysis as an acoustic classification task, we need to assign the whole audio segment into a single class and disregard the various temporal variations. To achieve this we set the pool size of the last max pooling layer so that its output averages all temporal information into a single temporal dimension. We tune the remaining convolutional block parameters, such as the number and size of filters used in each convolutional layer as well as the number of total blocks.
The output of the last convolutional block is then flattened, and mapped into the target number of classes with a final classification layer.

\subsubsection{Training procedure}
\begin{figure}
     \centering
     \begin{subfigure}[b]{\textwidth}
         \centering
         \includegraphics[width=0.8\textwidth]{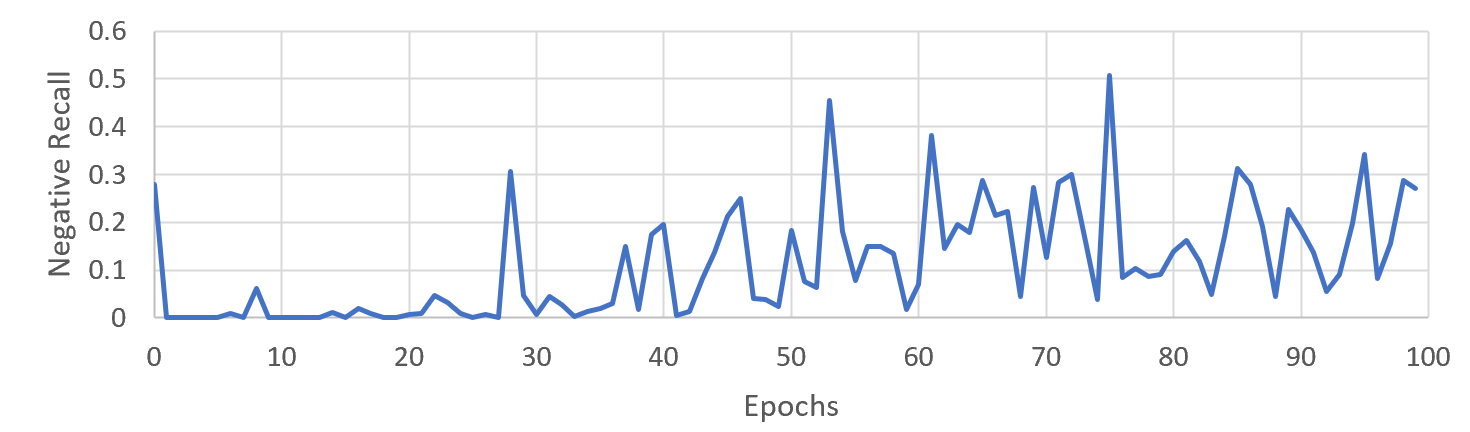}
         \caption{Without learning rate scheduler}
         \label{fig:no_scheduler}
     \end{subfigure}
     \hfill
     \begin{subfigure}[b]{\textwidth}
         \centering
         \includegraphics[width=0.8\textwidth]{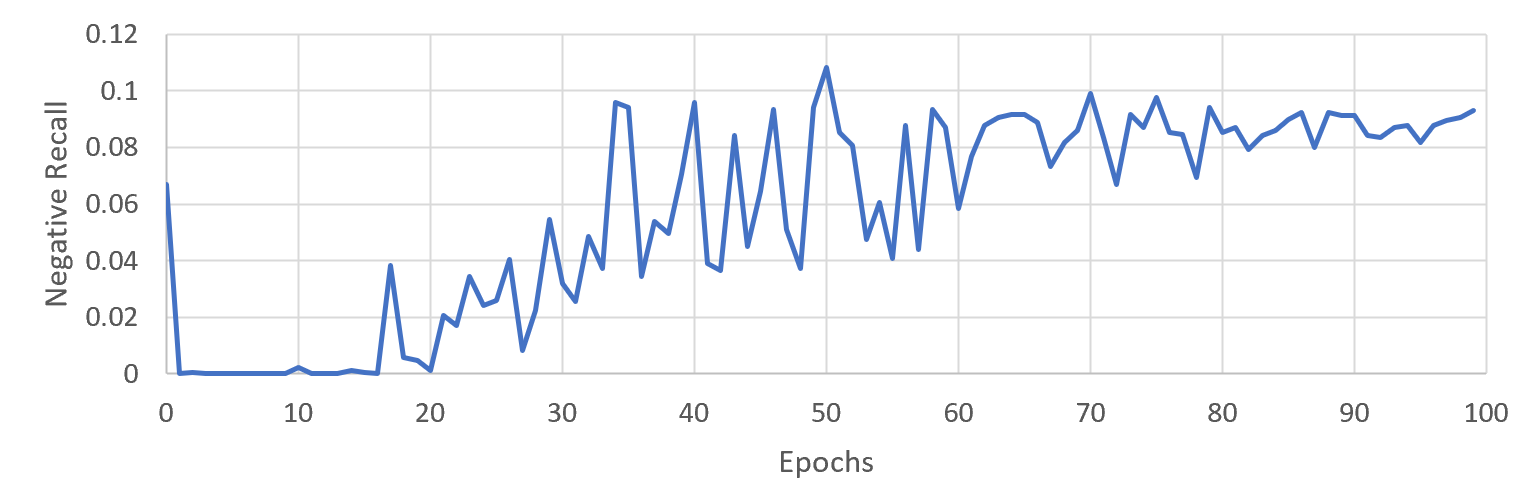}
         \caption{Using learning rate scheduler}
         \label{fig:scheduler}
     \end{subfigure}
     \hfill
        \caption{The recall in the negative class for the SWITCHBOARD validation set used during training.}
        \label{fig:neg_recall}
\end{figure}

The acoustic classification model was trained using the Adam optimiser \cite{kingma2014adam} and a \textit{sparse categorical cross entropy} loss function. 
During training we observed an erratic behavior in most of the validation metrics. 
This effect is shown in figure \ref{fig:no_scheduler}, where we can also observe that, as training progresses, we get random-like  negative class recall values, rapidly changing in the range 0 to 0.4. This indicates that any result is unstable and cannot be generalized. 

To overcome this instability issue, we use a learning rate scheduler, which monitors a specific training criterion, and reduces the learning rate by a certain factor once learning stagnates. For our experiments we used 0.5 as a learning rate factor, which results in an controlled reduction in learning rate so that we achieve a good convergence, while we are able to avoid local maxima. 
Concerning the monitored criterion, we have experimented with the most commonly used metrics in the evaluation of the sentiment analysis task, as for example the weighted accuracy (WA), unweighted accuracy (UA) and the recall values per class. As expected the convergence of the monitored metrics presents the most improvements, although the effects are visible in the remaining metrics as well.
For the above example, we can see in figure \ref{fig:scheduler}, the effect that the learning rate scheduler has on the monitored metric.

\subsection{Text-based sentiment analysis}
For the text-based sentiment analysis, we first computed word embeddings for the textual annotations using pre-trained BERT models \cite{devlin2018bert}. The size of each embedding vector was 300 and they were directly used as inputs to the neural network model.
The model consisted of consecutive 1D convolutional layers, followed by LSTM layers. The convolutional layers downsampled the input by a factor equal to the stride. Moreover, the kernel size used is always bigger than the stride, meaning the entire inputs are used to compute the output of each layer. The overall effect of this is that model is able to identify useful information while discarding unimportant details, making the output sequences shorter, which in turn can help the LSTM layer in detecting longer patterns \cite{geron2019hands}. The text-based models were more stable, with regards to metrics like  negative class recall, compared to the acoustic classification, therefore we did not introduce a learning rate scheduler for the training.

\subsection{Combination}
For the combination of the acoustic and text-based results into a bi-modal model, we use a late feature fusion approach. Specifically, we create two feature vectors for each speech utterance, which are calculated in the output of the pre-final layers of the two classifiers (acoustic and text-based). The two output vectors for each utterance are then concatenated to form a single vector and classified using a random forest classifier. The number of estimators hyperparameter was varied from 100 to 1000.

\subsection{Results}
The results for the above classifiers and the investigated datasets are presented in this section. First, we discuss the results for the single-modality classifiers, \textit{i.e} text and acoustic, and then the results in the bi-modal approach. We discuss the behaviour of each classifier per class, and the effect that the use of different monitoring criteria during training has on them.
We report results for various metrics commonly used in classification tasks: weighted accuracy (WA), unweighted accuracy (UA),  negative class recall (Ng. R.),  positive class recall (Ps. R) and neutral class recall (Nt. R). WA represents the standard accuracy measure, which is the total number of correct predictions divided by the total number of predictions. UA represents the average accuracy for every class, in other words it is the average of the individual recalls for all classes.
Furthermore, we report results for various training stopping criteria in order to understand the way that each classification setup behaves in the different classes, and identify the best approach to optimize the results in the underrepresented classes. 

\subsubsection{Text classification}
The fine-tuning process for the text-based classification model resulted in three convolutional layers of increasing filter sizes 32, 64, and 128, a kernel size 4 and stride 2, which were followed by an LSTM layer with 128 neurons. These parameters were used for both the SWITCHBOARD and IEMOCAP datasets as the tuning per dataset resulted in very similar setups.

\begin{table}
\centering
\caption{The results for the two investigated datasets in the text sentiment classification. For each row we monitor a specific metric.}
\label{tab:text_classification}
\vspace{1pt}
\begin{tabular}{l|ccccc|ccccc}
\centering
 & \multicolumn{5}{c|}{SWITCHBOARD}          & \multicolumn{5}{c}{IEMOCAP}        \\ \hline
\textit{\textbf{Monitoring Criteria}}  & WA  & Ng.R  & Ps.R  & Nt.R  & UA       & WA   & Ng.R  & Ps.R  & Nt.R  & UA \\ \hline
\textit{\textbf{WA}}    & 66.0  & 34.6  & 60.7  & 79.5   & 58.3 & 64.4 & 81.7 & 56.0  & 31.3 &   56.3 \\
\textit{\textbf{Ng.R}}  & 64.7 & 45.9  & 56.0  & 76.1  & 59.4 & 56.0  &97.8 & 13.5 & 0.0     &   37.1 \\
\textit{\textbf{Ps.R}} & 63.8 & 32.4  & 67.3  & 72.1  & 57.3     & 60.8 & 76.5 & 65.2 & 17.6 &    53.1\\
\textit{\textbf{Nt.R}}  & 62.9 & 22.2 & 39.8 &90.2 & 50.7 & 62.0  & 74.7 & 53.5 & 40.2 &    56.2 \\
\textit{\textbf{UA}}    & 65.1 & 38.7 & 62.2 & 75.6 &  58.8        & 63.8 & 78.4 & 59.6 & 32.8 &  56.9 \\
\hline
\end{tabular}
\end{table}

In table \ref{tab:text_classification}, we show the results for the text based sentiment classification.
For each reported monitoring case, the training runs for a variable number of epochs and stops when the corresponding metric stops improving.
First, we observe the difficulty of the classifier in accurately detecting the negative and neutral classes in the SWITCHBOARD and IEMOCAP datasets, which are the smaller classes respectively.
In both cases, monitoring the recall value of the smaller class, or directly the UA value, are good methodologies in order to improve the overall UA of the classifier.

\subsubsection{Acoustic classification} 
After the tuning process, the final acoustic model architecture for SWITCHBOARD consisted of three convolutional blocks with 64, 32 and 30 filters, followed by two fully connected layers of 128 and 64 neurons. The final IEMOCAP acoustic model architecture consisted of one convolutional block with 32 filters, followed by 1 fully connected layers with 32 neurons. The fact that the tuned IEMOCAP model is smaller than the respective SWITCHBOARD  model is expected, since the IEMOCAP dataset is significantly smaller than the SWITCHBOARD dataset.

\begin{table}
\centering
\caption{The results for the two investigated datasets in the acoustic sentiment classification. For each row we monitor a specific metric using a learning rate scheduler.}
\label{tab:acoustic_classification}
\vspace{1pt}
\begin{tabular}{l|ccccc|ccccc}
\centering
 & \multicolumn{5}{c|}{SWITCHBOARD}          & \multicolumn{5}{c}{IEMOCAP}        \\ \hline
\textit{\textbf{Monitoring criteria}}  & WA  & Ng.R  & Ps.R  & Nt.R  & UA    & WA   & Ng.R  & Ps.R  & Nt.R  & UA \\ \hline
\textit{\textbf{WA}}    & 57.1    & 0.0     & 24.7  & 95.2   &   40.0   & 61.2 & 90.6 & 25.5 & 28.1 &  48.1  \\
\textit{\textbf{Ng.R}}  & 51.5    & 19.5    & 34.9  & 71.9  &   42.1    & 56.6  & 99.0  & 5.7  & 8.6     &  37.8  \\
\textit{\textbf{Ps.R}} & 44.3    & 11.9    & 69.7  & 40.0  &  40.5     & 44.7 & 48.6 & 71.6 & 5.5 &    41.9\\
\textit{\textbf{Nt.R}}  & 56.9  & 0.0     & 25.2 &94.6 &   39.9   & 50.0  & 53.3  & 18.4 & 76.6 &   49.4 \\
\textit{\textbf{UA}}    & 48.4  & 29.5  & 53.1 & 51.8 &  44.8     & 59.1 & 78.1 & 25.5 & 49.6 &  51.1 \\
\hline
\end{tabular}
\end{table}

The results of the acoustic classification on both datasets are shown in table \ref{tab:acoustic_classification}.  As before, the monitoring setups concern the use of different stopping criteria. As described before, we also use an learning rate scheduler in matched conditions, meaning that the scheduler and the early stopping callbacks monitor the same value.
First, we observe that the acoustic classification results are greatly affected by the selection of the monitoring criterion. For example, in the SWITCHBOARD dataset the  negative class recall has increased from 0\% to 19\% by changing the monitoring from WA to Ng.R. 
As observed for text classification, selecting the UA offers a good trade-off among the classes. However, in a real application one should consider the expected behaviour of the system. In many cases, improving the prediction in the negative and positive classes is more critical than improving the UA.

\subsubsection{Bi-modal classification}
The tuning of the random forest hyperparameters produced similar results, therefore we report the results of the simple classifier with 100 estimators.
In table \ref{tab:bi_modal}, we report the results from the bi-modal classification. 
For the combination we use the output of the prefinal layers of the acoustic and text classification. The shapes of the acoustic vector are 64 and 32 for SWITCHBOARD and IEMOCAP respectively, and 128 for each text output. 
The different monitoring setups here concern the corresponding single modalities used. For instance, a   negative class recall monitoring means that both the text and the acoustic classifiers are trained using this stopping criterion.
Although the bi-modal classification does help in improving the overall classification metrics, WA and UA, it does not offer overall significant improvements. This may be attributed to the selection of the simple classifier, \textit{i. e} a random forest, as opposed to a more sophisticated neural architecture.

\begin{table}
\centering
\caption{The results for the two investigated datasets in bi-modal sentiment classification. For each row we monitor a specific metric.}
\label{tab:bi_modal}
\vspace{1pt}
\begin{tabular}{l|ccccc|ccccc}
\centering
 & \multicolumn{5}{c|}{SWITCHBOARD}          & \multicolumn{5}{c}{IEMOCAP}        \\ \hline
\textit{\textbf{Monitoring criteria}}  & WA  & Ng.R  & Ps.R  & Nt.R  & UA       & WA   & Ng.R  & Ps.R  & Nt.R  & UA \\ \hline
\textit{\textbf{WA}}    & 67.0    & 39.0  & 61.0  & 79.0   & 60.0     & 67.5 & 79.0 & 62.0 & 44.0 &   62.0 \\
\textit{\textbf{Ng.R}}  & 61.0    & 30.0  & 52.0  & 76.0  &  53.0     & 64.0  & 79.0  & 50.0  & 43.0    & 58.0   \\
\textit{\textbf{Ps.R}} & 65.0    & 38.0  & 57.0  & 79.0  &  58.0     & 68.0 & 82.0 & 56.0 & 48.0 &  62.0  \\
\textit{\textbf{Nt.R}}  & 66.0  & 34.0    & 59.0  & 81.0 &   58.0    & 63.0  & 76.0  & 57.0 & 41.0 & 58.0  \\
\textit{\textbf{UA}}    & 67.0  & 39.0    & 61.0  & 79.0 &   60.0     & 68.0 & 79.0 & 63.0 & 46.0 &  63.0 \\
\hline
\end{tabular}
\end{table}

It is interesting though to point out that in the IEMOCAP case, the bi-modal classifier is good at combining complementary information from the two modalities. Notice the improvement that is reported in the underrepresented classes (positive and neutral) while maintaining a good recall in the negative class, when compared with the corresponding results coming from the text modality.
The same however is not evident in SWITCHBOARD, something that can be explained by the nature of the dataset. SWITCHBOARD is a dataset built for automatic speech recognition and the conversations regard topics given as a prompt to the speakers. The speakers were not instructed to include any emotional state in their speech and the contents do not cause strong emotions, therefore we expect less sentimental acoustic cues for utterances that are deemed negative or positive. In the positive class, phonations such as laughs may help the model make more accurate predictions. 

Nevertheless, the SWITCHBOARD results compare favourably to the results reported in \cite{chen2020large},  as shown in table \ref{tab:comparison}.
There are no comparable results for 3-class IEMOCAP experiments since most experiments involving this dataset concern emotion recognition and use more classes from the IEMOCAP label set.

\begin{table}
\caption{Comparative results for the SWITCHBOARD dataset}
\label{tab:comparison}
\centering
\begin{tabular}{c|c|c|c}
                            Modality      &   Metric    & Chen et al. \cite{chen2020large} & Proposed solution \\
                            \hline
\multirow{5}{*}{Acoustic} & WA    & 54.2 &     48.4          \\
                          & UA    & 39.6 &     44.8              \\
                          & Ng. R & 0.0     &     29.5          \\
                          & Ps. R & 40.0    &     53.1              \\
                          & Nt. R & 78.0    &     51.8              \\ \hline
\multirow{5}{*}{Bi-modal}         & WA    & 65.6 &         67.0          \\
                                              & UA    & 54.6 &  60.0                  \\
                                              & Ng. R &   --    &   39.0                 \\
                                              & Ps. R &--  &    61.0               \\
                                              & Nt. R &  --     & 79.0                 
\end{tabular}
\end{table}

\section{Conclusions} \label{sec:conclusions}
In this work, we demonstrated that bi-modal analysis perform better when the data have an inherent sentiment component to them, as opposed to an everyday conversational nature. 
Therefore, according to the nature of each dataset, different configurations may lead to optimal results. For example, in the SWITCHBOARD dataset where sentiment is not expected to be emphasized in oral speech we observed that text is a good source for sentiment analysis. In IEMOCAP where we generally expect strong acted sentiment we observe that the acoustic information becomes much more relevant. 
In such datasets, bi-modal models are successful in detecting sentiment, also in underrepresented classes using fusion techniques.
This conclusion however, indicates the general need for more realistic sentiment analysis corpora for building bi-modal solutions.

In general, we have observed that, as expected, classification accuracy suffers in the smaller classes, which for sentiment analysis are often the non-neutral classes. To address this we introduced a monitoring criterion which focuses on the problematic classes, reduces learning rate in the audio classification, and also selects the best stopping point during training. In any case, the combination strategy that we implement is able to benefit from both information sources, acoustic and text, and significantly improve the results for all sentiment classes.

Finally, to the best of our knowledge, the bi-modal results reported for the SWITCHBOARD dataset are the best achieved from comparable approaches in the literature. In \cite{chen2020large}, the use of pretrained ASR features is reported to further improve the results, however not in the negative class. Also one needs to take into account the increase in the resources that is required for the training and subsequent use of a sufficiently large pretrained ASR system.
\subsubsection*{Acknowledgements}
This work has received funding from the European Union’s Horizon 2020 research and innovation program under grant agreement No 872139, project aiD. 
We would like to thank our colleagues Steve Barrett, Zacharias Georgiou and Andrey Filyanin for their valuable help and feedback in the preparation of this work.

%
%

\bibliographystyle{splncs04}
\bibliography{bibliography}

\end{document}